\documentclass{article}

\usepackage{arxiv}

\usepackage[utf8]{inputenc} 
\usepackage[T1]{fontenc}    
\usepackage{hyperref}       
\usepackage{url}            
\usepackage{booktabs}       
\usepackage{amsfonts}       
\usepackage{nicefrac}       
\usepackage{microtype}      
\usepackage{lipsum}
\usepackage{graphicx}
\graphicspath{ {./images/} }

\usepackage{amsmath}
\usepackage{algorithm}
\usepackage{algorithmic}
\usepackage{multirow}  
\usepackage{booktabs}  
\usepackage{xcolor}

\title{Revealing the Challenges of Sim-to-Real Transfer in Model-Based Reinforcement Learning \\ via Latent Space Modeling }

\author{
  Zhilin Lin \quad Shiliang Sun \\
  School of Automation and Intelligent Sensing \\ Shanghai Jiao Tong University \\
  800 Dongchuan Road, Shanghai 200240, P.R. China \\
  \texttt{linzl123@sjtu.edu.cn} \quad \texttt{shiliangsun@gmail.com}
}

\begin{document}
\maketitle
\begin{abstract}
Reinforcement learning (RL) is playing an increasingly important role in fields such as robotic control and autonomous driving. However, the gap between simulation and the real environment remains a major obstacle to the practical deployment of RL. Agents trained in simulators often struggle to maintain performance when transferred to real-world physical environments. In this paper, we propose a latent space based approach to analyze the impact of simulation on real-world policy improvement in model-based settings. As a natural extension of model-based methods, our approach enables an intuitive observation of the challenges faced by model-based methods in sim-to-real transfer. Experiments conducted in the MuJoCo environment evaluate the performance of our method in both measuring and mitigating the sim-to-real gap. The experiments also highlight the various challenges that remain in overcoming the sim-to-real gap, especially for model-based methods.
\end{abstract}


\section{Introduction}
In real-world applications of reinforcement learning (RL), there are typically two types of environments involved. The first is the \textbf{real environment}, which represents the actual deployment setting where the policy is ultimately expected to perform. Achieving strong performance in the real environment is the final goal of training; only policies that succeed under real-world conditions can be considered truly effective. However, interactions in the real environment are often costly and limited due to practical constraints. The second is the \textbf{simulation}, which is introduced to compensate for the lack of interaction data from the real world. While simulators offer the advantage of virtually unlimited, low-cost interactions, they often fail to perfectly replicate the dynamics of real environments. As a result, policies trained solely in the simulation may not generalize well or perform reliably when deployed in the real world.

Effectively leveraging the simulation’s scalability alongside the dynamic fidelity of the real environment to improve RL policy performance in real-world settings has become a central research challenge. Coordinated utilization of both environments is a key to accelerating the practical deployment of RL technologies. In practical scenarios, this coordination problem between simulation and real environment has emerged as a core bottleneck. While the real environment is the ultimate testbed for policy validation, its high interaction cost and uncontrollable risks significantly hinder training efficiency. Take autonomous driving as an example: although Waymo has accumulated millions of miles of real-world driving data, its limited exposure to long-tail scenarios such as extreme weather or unexpected pedestrian behavior continues to hinder commercialization~\cite{schwall2020waymo}. A similar challenge exists in industrial robotics, where modeling errors—such as those in contact dynamics between robotic arms and deformable objects—can lead to policy failure. This challenge has fueled the widespread use of simulation environments, which offer low-cost interaction opportunities to address real-world data scarcity. However, discrepancies in simulation modeling still pose major barriers to policy transfer. For instance, while the CARLA simulator can render highly detailed driving scenes, it struggles to fully capture the dynamic variations in tire-road friction coefficients, leading to unstable policy performance on slippery surfaces~\cite{dosovitskiy2017carla}.

So, determining when to trust the simulator and how to effectively leverage it to enhance agent performance in the real world are key challenges. To address the above two challenges, some methods have been proposed in recent years, most of which are model-free methods. Niu et al.~\cite{niu2022trust} introduces a method that applies a degree of pessimistic estimation to the Q-values of states in the simulation, based on the discrepancy between samples from the simulation and those from the real world. Furthermore, it leverages importance sampling to reweight simulated samples by learning a mapping between the state transitions in the two environments, thereby aligning the simulated data distribution with that of the real environment. Although this approach significantly enhances the usability of simulated samples, it does not propose a unified metric for quantifying the dynamic discrepancy between the two environments, and its use of simulation data remains relatively limited. Subsequent work~\cite{niu2023h2o+, hou2024improving} further refine the use of simulated samples. However, they continue to focus on learning sample-level mappings from simulation to real environment to compensate for the lack of real data, without delving deeper into the fundamental similarities and differences between the two environments.

Meanwhile, studies on model-based methods for sim-to-real transfer are relatively scarce, and challenges faced by these methods remain underexplored. Model-based methods, which explicitly learn environment dynamics for planning or policy improvement, are particularly sensitive to the sim-to-real gap. Despite their high sample efficiency in simulation, these methods often suffer from model inaccuracies and overfitting to simulator-specific dynamics, which significantly hinder their real-world applicability. Understanding the specific ways in which the sim-to-real gap impacts model-based reinforcement learning remains an open and important research question.

Latent space methods have shown strong potential in transfer learning, particularly due to their capacity to model shared structures across domains. Building on this idea, we propose a latent space based approach not only as a sim-to-real transfer method, but also as an analytical tool to better understand and evaluate the challenges faced by model-based reinforcement learning under sim-to-real gap. In our framework, the latent space is designed to capture both the connections and discrepancies between simulation and real environment. It not only encodes invariant dynamic patterns across domains, but also exposes where and how learned representations diverge—thereby offering insights into the root causes of policy degradation in sim-to-real transfer. As a tool for discovering the challenges faced by model-based methods, this approach can be naturally integrated into various algorithms and has high flexibility.

By embedding observations into a latent space, we effectively augment single-environment observations with additional dynamic information, potentially transforming Partially Observable Markov Decision Processes (POMDPs) into fully observable ones. Furthermore, during training, the latent space can autonomously discover relationships between states and transitions across environments. The mapping functions learned in this space serve as implicit indicators of the dynamic sim-to-real gap, providing meaningful guidance for the utilization of simulation data.

In this paper, we propose a latent space based reinforcement learning algorithm to both quantify and mitigate the sim-to-real gap, and use this approach to uncover the challenges it poses for real-world policy training. Our main contributions are summarized as follows:
\begin{itemize}

\item We introduce a latent space based method that learns to estimate the magnitude of the sim-to-real gap during training. This serves as a quantitative metric to assess how simulation impact real-world performance.

\item Our proposed latent space based methods can discover the differences and connections between environmental observations by exploring deeper features with different dynamic environments, thereby mitigating the negative effects of the sim-to-real gap on policy learning.

\item Through preliminary experiments and analysis, we identify and summarize key challenges introduced by the sim-to-real gap in practical policy training, and outline potential future research directions for addressing them.
\end{itemize}

\section{Related Works}

\subsection{Transfer Reinforcement Learning}
Transfer Reinforcement Learning (TRL) has emerged as a prominent research direction in machine learning, driven by the practical limitations of conventional RL. In standard RL frameworks, agents learn optimal policies through trial-and-error interactions with the environment, which often require a vast number of training samples and long convergence times. These limitations are especially pronounced in complex dynamic settings, where high training costs and weak policy generalization remain persistent challenges~\cite{sutton1998reinforcement}.

To address these issues, researchers have proposed the use of \textbf{knowledge transfer mechanisms} to enable experience reuse across tasks, giving rise to the field of TRL. At its core, TRL aims to establish a systematic paradigm for transferring knowledge—such as policy networks, value functions, or environment models—learned in a source task to accelerate learning in a target task. This is achieved through methods such as feature mapping, parameter sharing, or experience reuse, significantly improving exploration efficiency in new environments~\cite{taylor2009transfer}.

Based on differences in MDP components, existing TRL approaches are commonly categorized into three types. The first is state-action heterogeneity, where the observation or action spaces differ between tasks. The second is dynamics heterogeneity, which involves shifts in the transition probability $P(s'|s,a)$ or the reward function $R(s,a)$ and sim-to-real transfer is a special case of this type. The third is compound heterogeneity, where both the observation space and environment dynamics differ simultaneously, this is the hardest type.

Methodologically, research in TRL follows three major directions. The first is reward shaping, which integrates domain knowledge into the reward function design. This includes the use of potential-based shaping functions to guide agents toward critical states~\cite{ng1999policy}, or curriculum learning strategies that progressively adjust reward signals~\cite{bengio2009curriculum}. The second is policy transfer, which focuses on combining and adapting multiple source policies. Techniques such as policy distillation can aggregate expert knowledge into a unified transferable policy~\cite{rusu2015policy}, while meta-transfer algorithms enable quick adaptation using few target domain samples~\cite{finn2017one}. The third is representation alignment, which aims to build shared feature spaces across domains. For example, deep correlation networks can extract domain-invariant state representations~\cite{ganin2016domain}, and adversarial transfer frameworks can align dynamic behavior through domain discriminators~\cite{tzeng2017adversarial}. These techniques have demonstrated strong performance across a variety of real-world applications, including adaptive parameter tuning for industrial robots~\cite{peng2018sim}, scenario adaptation in autonomous driving~\cite{sallab2017deep}, and cross-domain learning in medical image diagnosis~\cite{esteva2021deep}.

\subsection{Model-Based Reinforcement Learning}

Sample efficiency is one of the key challenges in RL, especially in deep reinforcement learning (DRL)~\cite{mnih2015human}. The learning of RL policies relies heavily on interaction data between agents and environments. When deploying RL algorithms in real-world scenarios, the high cost of exploration and the difficulty of interaction make enhancing sample efficiency particularly critical. To address this issue, model-based reinforcement learning (MBRL) has been proposed. These methods aim to build a model of the environment using limited interaction data, enabling the agent to generate additional training samples through simulated rollouts, thereby significantly reducing the burden of data collection for policy learning.

Early approaches, such as Dyna-Q~\cite{sutton1990integrated}, integrated planning and learning by leveraging simulated experiences, showing promising results in environments with discrete state spaces. A major challenge in MBRL is the accumulation of errors due to imperfect environment models. To mitigate this, probabilistic models were introduced to capture uncertainty in environment dynamics, as exemplified by methods like PILCO~\cite{deisenroth2011pilco} and PETS~\cite{chua2018deep}. More recent approaches, such as MBPO~\cite{janner2019trust}, improve the effectiveness of model usage in policy optimization through the "branched rollout" technique. Its extensions~\cite{yu2020mopo, rigter2022rambo} are specifically designed to address challenges in offline settings~\cite{huang2024pessimism}.

Recent advances in model-based reinforcement learning, such as the Dreamer series~\cite{hafner2019dream, hafner2020mastering, hafner2023mastering}, have demonstrated the effectiveness of learning latent dynamics models from high-dimensional pixel observations. These methods encode raw visual inputs into compact latent representations, facilitating efficient policy learning in visually complex environments. While our work is not focused on pixel-based tasks, it draws inspiration from the idea of latent modeling to investigate the underlying structure of dynamics across simulation and real environment. In contrast to Dreamer's end-to-end policy learning in latent space, our method uses the latent representation as an analytical tool to better understand, measure, and potentially mitigate the sim-to-real gap. By integrating this latent modeling framework with the MBPO algorithm, we develop a model-based approach tailored for analyzing policy transfer challenges in sim-to-real scenarios.

\subsection{Reinforcement Learning Mitigating Sim-to-Real Gap}
Due to the imperfections of simulators, policies trained in simulation often fail to transfer directly to the real environment. As a result, a number of reinforcement learning methods have been proposed to mitigate the sim-to-real gap. One common approach is domain randomization~\cite{peng2018sim, mehta2020active}, which involves randomizing the configuration of simulations to enhance the robustness of the learned policies. Other methods focus on directly modifying the transition dynamics of the simulator using data from the real environment. For example, Chebotar et al.~\cite{chebotar2019closing} optimize the simulator's hyperparameters by matching trajectories from both simulation and real environment to achieve similar dynamics. Liang et al.~\cite{liang2025novel} learn a mapping between simulation and real environment and use it to directly adjust the interaction data in the simulator. Jiang et al.~\cite{jiang2018pac} iteratively refine the transition probabilities in the simulator based on real-world transitions, aiming to align the simulator more closely with reality, and provide theoretical analysis under a probably approximately correct (PAC) framework.

The H2O series of works~\cite{niu2022trust, niu2023h2o+, hou2024improving} adopt a hybrid offline-and-online training scheme, where offline data from the real environment and online data from simulation are jointly utilized. By introducing a modified Bellman error and incorporating pessimism into the value estimation, they improve the utility of simulation-generated data, thereby enhancing policy performance in the real world. However, these approaches are fundamentally model-free and do not explicitly model environment dynamics or domain differences.

In contrast, our method operates under the same hybrid training setting but introduces a model-based framework with a shared latent space to analyze and quantify the sim-to-real gap. By learning structured latent representations of both environments, our approach explicitly captures dynamic similarities and discrepancies, providing a principled way to diagnose the impact of the sim-to-real gap on policy learning. This analytical capability distinguishes our method from existing approaches, offering new insights into the limitations of model-based reinforcement learning under domain shifts.

\section{Preliminaries}

\subsection{Reinforcement Learning}
We consider a standard RL setting~\cite{sutton1998reinforcement,sun2020pattern}, where an agent interacts with an environment modeled as a Markov Decision Process (MDP), defined by the tuple $(\mathcal{S}, \mathcal{A}, P, R, \gamma)$. Here, $\mathcal{S}$ is the state space, $\mathcal{A}$ is the action space, $P(s'|s,a)$ denotes the transition probability from state $s$ to $s'$ given action $a$, $R(s,a)$ is the reward function, and $\gamma \in [0,1)$ is the discount factor. The agent learns a policy $\pi(a|s)$, which maps states to a distribution over actions, with the objective of maximizing the expected cumulative discounted reward:
\begin{equation}
J(\pi) = \mathbb{E}_{\pi}\left[\sum_{t=0}^{\infty} \gamma^t r(s_t, a_t)\right].
\end{equation}

In practical applications, especially in robotics, RL policies are often trained in simulations. In this case, the agent faces two different MDP environments, one corresponding to the real environment and the other corresponding to the simulation, which we denote as $\mathcal{M}$ and $\mathcal{M^{\prime}}$ respectively:
\begin{equation}\mathcal{M}: =(\mathcal{S}, \mathcal{A}, P_{\mathcal{M}}, R_{\mathcal{M}}, \gamma),\, \mathcal{M^{\prime}}: =(\mathcal{S}, \mathcal{A},  P_{\mathcal{M^{\prime}}}, R_{\mathcal{M^{\prime}}}, \gamma).\end{equation}

Two environments share the same state space, action space and discount factor, but differ in their reward functions or transition dynamics. Formally, there exist $(s, a) \in \mathcal{S} \times \mathcal{A}$ and $s' \in \mathcal{S}$ such that
\[
R_{\mathcal{M'}}(s,a) \neq R_{\mathcal{M}}(s,a) \quad \text{or} \quad P_{\mathcal{M'}}(s' | s,a) \neq P_{\mathcal{M}}(s' | s,a).
\]

The objective is to optimize the expected cumulative discounted return in the real environment, that is, to find a policy $\pi$ that maximizes
\begin{equation}
J(\pi) :=\mathbb{E}_{a_t\sim\pi(\cdot | s_t),\; s_{t+1}\sim P_{\mathcal{M}}(\cdot|s_t, a_t)} \left[\sum_{t=0}^\infty \gamma^t R_{\mathcal{M}}(s_t, a_t)\right].
\end{equation}

The main challenge lies in the fact that the environment used for training the policy $\pi$ (i.e., $\mathcal{M'}$) does not perfectly match the target environment $\mathcal{M}$ in terms of dynamics and rewards, which makes it difficult to ensure the effectiveness of the learned policy after transfer.

\subsection{Model-Based Policy Optimization (MBPO)}

Model-Based Policy Optimization (MBPO) is a RL algorithm that combines the sample efficiency of model-based methods with the high performance of model-free approaches. MBPO reduces error accumulation by carefully limiting the horizon of model-generated rollouts and interleaving them with real environment interactions.

Specifically, MBPO leverages an ensemble of probabilistic dynamics models to learn the transition dynamics of the environment from real-world data. These learned models are then used to generate synthetic data by performing short-horizon rollouts starting from states sampled from a real replay buffer. This synthetic data is combined with real experience to train a model-free RL algorithm, such as Soft Actor-Critic (SAC)~\cite{haarnoja2018soft}, which acts as the policy optimization backbone. By dynamically adjusting the rollout horizon and using model uncertainty to guide policy updates, MBPO maintains stability and mitigates the compounding errors typically associated with long-horizon model rollouts.

MBPO has demonstrated strong empirical performance across standard continuous control benchmarks, offering a favorable trade-off between sample efficiency and asymptotic performance. More importantly, it is the basis for many model-based methods.
In this work, our use of environment model is mainly designed based on the MBPO algorithm to improve the reliability of environment model in policy training.

\section{Method}

In this section, we will introduce our latent space based approach. \ref{4.1} introduces the detailed design ideas of the latent space and explains why our proposed latent space has the ability to discover the gaps and connections between different environments. \ref{4.2} introduces how to use the latent space training policy. \ref{4.3} introduces a practical algorithm and summarizes this section.

\begin{figure} 
    \centering
    \includegraphics[width=1\linewidth]{  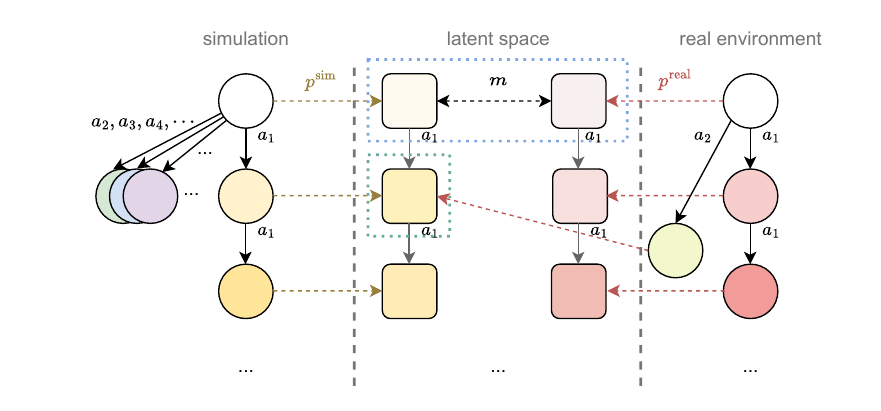}
    \caption{Latent state based environment modeling.
$\bigcirc$ shows the observation in the simulation and real environment,
\includegraphics[height=1em]{  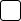} shows the state in the latent space, \includegraphics[height=1em]{  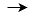} denotes the transition when taking an action at a given state.
\includegraphics[height=1em]{  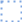} indicates that the same observation in simulation and real environments is mapped to \textbf{different latent states} via
\includegraphics[height=1.2em]{  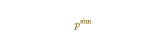} and
\includegraphics[height=1.2em]{  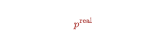}.
\includegraphics[height=1em]{  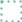} shows that different observations in simulation and real environments may be mapped to \textbf{close or even the same latent state}, facilitating a unified analysis of dynamics across domains.}
    \label{fig:method}
\end{figure}

\subsection{Latent Space Design}\label{4.1}

We propose a latent space based environment modeling framework to address the mismatch in dynamics between simulation and real environments. As shown in Figure~\ref{fig:method}, this discrepancy often arises because the observable state $s$ does not capture all critical variables influencing dynamics, resulting in a partially observable process. While both environments can individually be modeled as MDPs, they together form a Partially Observable MDP (POMDP). A state in a single environment is merely an observation in the joint environment.

To recover full observability, an intuitive approach is to augment the state with an additional variable representing essential but hidden factors. For instance, a robot walking on snow and dry ground may receive the same state $s$ but experience different transitions due to friction differences. By embedding such environmental characteristics in a new state $\bar{s} = [s; e]$, we achieve a fully observable formulation, where $e$ represents inferred environment features.

Learning environment features often needs to be done in multiple environments, and is not suitable for direct use in sim-to-real tasks. However, inspired by this idea, we can design mappings from state in each environment to a common latent space, where the model can automatically learn the dynamic differences between environments. We hope that the states in the latent space can contain more dynamic information to approach the fully observed MDP. For the same observed state (i.e., observation for the joint environment) in the simulation and the real environment, we distinguish them in the latent space because of their different transition dynamics; and for different observations in the two environments, if they have some similarity in dynamic characteristics, we hope that their corresponding states in the latent space are closer. We define a latent space MDP as:
\begin{equation}
\bar{\mathcal{M}} := (\bar{\mathcal{S}}, \mathcal{A}, P_{\bar{\mathcal{M}}}, R_{\bar{\mathcal{M}}}, \bar{\rho}, \gamma),
\end{equation}
where $\bar{\mathcal{S}}$ is the latent state space shared by both simulation and real environment, $\mathcal{A}$ is the common action space, and $\gamma$ the discount factor. 

We introduce mapping functions $p^{\text{sim}}, p^{\text{real}}: \mathcal{S} \rightarrow \bar{\mathcal{S}}$ to project observable states from each environment into the latent space. Additionally, we define a cross-domain mapping $m: \text{Image}(p^{\text{real}}) \rightarrow \text{Image}(p^{\text{sim}})$ such that for any $o \in \mathcal{S}$,
\[
m \circ p^{\text{real}}(o) \approx p^{\text{sim}}(o),
\]
providing a quantitative measure of the environment gap.

To learn an effective latent space, we jointly optimize for model prediction accuracy, auto-encoding fidelity, and cross-domain alignment. The objectives are as follows:

\paragraph{1. Model Prediction Objectives:}
\begin{align}
&\|P_{\bar{\mathcal{M}}} (\cdot| p^{\text{sim}}(s), a) - P_{\mathcal{M^\prime}}(\cdot| s, a)\|_{\text{TV}}, \\
&\|P_{\bar{\mathcal{M}}} (\cdot| p^{\text{real}}(s), a) - P_{\mathcal{M}}(\cdot| s, a)\|_{\text{TV}}, \\
&\|R_{\bar{\mathcal{M}}} (p^{\text{sim}}(s), a) - R_{\mathcal{M^\prime}}(s, a)\|_{2}, \\
&\|R_{\bar{\mathcal{M}}} (p^{\text{real}}(s), a) - R_{\mathcal{M}}(s, a)\|_{2}.
\end{align}

\paragraph{2. Auto-Encoding Objectives:}
Let $q^{\text{sim}}, q^{\text{real}}$ be decoders for $p^{\text{sim}}, p^{\text{real}}$, respectively:
\begin{align}
\mathbb{E}_{D_\text{sim}}[\|s - q^{\text{sim}} \circ p^{\text{sim}}(s)\|_2 + \|s' - q^{\text{sim}} \circ p^{\text{sim}}(s')\|_2], \\
\mathbb{E}_{D_\text{real}}[\|s - q^{\text{real}} \circ p^{\text{real}}(s)\|_2 + \|s' - q^{\text{real}} \circ p^{\text{real}}(s')\|_2].
\end{align}

\paragraph{3. Latent Correspondence Objectives:}
\begin{equation}
\mathbb{E}_{D_\text{sim} \cup D_\text{real}}[\|p^{\text{sim}}(s) - m \circ p^{\text{real}}(s)\|_2 + \|p^{\text{sim}}(s') - m \circ p^{\text{real}}(s')\|_2].
\end{equation}

We parameterize $m$ as a separate lightweight neural network, instead of using $p^{\text{sim}} \circ q^{\text{real}}$, to enhance training stability and interpretability.

To reduce sample complexity in the real world, we consider two strategies to initialize the latent dynamics model: using simulated data or using real-world offline data. We begin by training the environment model within a single environment using its corresponding data. When data from another environment is introduced, we initialize the encoder for the current environment as an identity mapping—i.e., $p^{\text{sim}} = \text{id}$ for simulation-based initialization, or $p^{\text{real}} = \text{id}$ for real-environment-based initialization. Accordingly, we define:

\[
P_{\bar{\mathcal{M}}}|_\mathcal{S} := P_{\mathcal{M}^\prime}, \quad R_{\bar{\mathcal{M}}}|_\mathcal{S} := R_{\mathcal{M}^\prime} \quad \text{(simulation initiation)},
\]

or

\[
P_{\bar{\mathcal{M}}}|_\mathcal{S} := P_{\mathcal{M}}, \quad R_{\bar{\mathcal{M}}}|_\mathcal{S} := R_{\mathcal{M}} \quad \text{(real-environment initiation)}.
\]

After the initialization, both real-world offline data and simulated data are jointly used to train the latent models. Real-world data improves model fidelity in critical regions of the state space, while simulated rollouts help guide model optimization in alignment with the policy improvement direction.

This design enables policies pre-trained in a single environment to be seamlessly initialized in the latent space and adapted using mixed-domain data. As a result, knowledge acquired from simulation is effectively reused, and the latent model continuously evolves to bridge the sim-to-real gap.

\subsection{Latent Space Utilization in Policy Learning}\label{4.2}

Once the latent model is trained, policy optimization is conducted in the latent state space $\bar{\mathcal{S}}$, where both simulation and real-world transitions can be modeled consistently. The latent space return objective is:
\begin{equation}
\bar{J}(\pi) := \mathbb{E}_{a_t \sim \pi(\cdot|\bar{s}_t), \bar{s}_{t+1} \sim P_{\bar{\mathcal{M}}}(\cdot|\bar{s}_t, a_t)} \left[ \sum_{t=0}^\infty \gamma^t R_{\bar{\mathcal{M}}}(\bar{s}_t, a_t) \right].
\end{equation}
In the ideal case where the latent model perfectly matches both the simulation and real environment, optimizing $\bar{J}(\pi)$ is equivalent to maximizing the true return in both environment, up to a change in initial state distribution. We use MBPO to optimize the policy in the latent space based on the environment model for the latent space.

In the evaluation phase of the policy, since the policy is executed in the latent space, when executing each action, the states in the simulation and real environment need to be mapped to $\bar{\mathcal{S}}$ through $p^{\text{sim}}$ and $p^{\text{real}}$ and then input into the trained policy model. If the latent space is well trained and has the ability to make good distinctions and discover associations between the states of different environments, then the performance in both the simulation and real environment is expected to be improved simultaneously.

\subsection{Overall Algorithm}\label{4.3}

\begin{algorithm}[H]
\caption{Latent Space Based Reinforcement Learning Policy Optimization}
\label{alg1}
\begin{algorithmic}[1]
\STATE Train the state transition function $p_\theta$, reward function $r_\theta$, and current policy $\pi_\phi$ using either interaction data from the simulation or offline data $\mathcal{D}_{\text{offline}}$ from the real environment.
\STATE Initialize the latent space state transition function $P_{\bar{\mathcal{M}}}$ and reward function $R_{\bar{\mathcal{M}}}$ using the models learned in Step 1.
\STATE Initialize the interaction dataset $\mathcal{D}_{\text{env}}$ in the simulation.
\FOR{$N$ iterations}
    \STATE Update $P_{\bar{\mathcal{M}}}$, $R_{\bar{\mathcal{M}}}$, $p^\text{sim}$, $p^\text{real}$, and $m$ in the latent space using samples from $\mathcal{D}_{\text{env}}$ and $\mathcal{D}_{\text{offline}}$, according to the optimization objective in \ref{4.1}.
    \FOR{$E$ steps of interaction in the simulation}
        \STATE Collect samples using the current policy $\pi_\phi$ in the simulation and add them to $\mathcal{D}_{\text{env}}$.
        \FOR{$M$ times of latent space model rollout}
            \STATE Uniformly sample $s_t$ from $\mathcal{D}_{\text{env}}$.
            \STATE From $s_t$, perform $k$-step rollouts in latent space using policy $\pi_\phi$ and add the data to $\mathcal{D}_{\text{model}}$.
        \ENDFOR
        \FOR{$G$ gradient updates}
            \STATE Optimize the policy on $\mathcal{D}_{\text{model}}$ using SAC.
        \ENDFOR
    \ENDFOR
\ENDFOR
\end{algorithmic}
\end{algorithm}

The pseudocode of our method is shown in \ref{alg1}. Our method broadly follows the framework of MBPO, but introduces targeted modifications to the algorithmic process due to the distinction between the simulation and real environments. More importantly, our latent space based algorithm is designed primarily as a tool to analyze the challenges faced by model-based methods in sim-to-real transfer, by explicitly modeling and quantifying the discrepancies and connections between the two environments within the latent space.

Specifically, unlike the original MBPO algorithm, we do not begin training with randomly initialized environment models and policies. Instead, we first train a preliminary environment model and policy using a single environment (either the simulation or the real one), and use them to initialize the models and policy in the latent space. This initialization strategy provides a better starting point for policy learning within the latent space.

Subsequently, in each iteration, we train the environment model in the latent space rather than in a single environment. During this phase, we adopt the optimization objective introduced in \ref{4.1}. We then adopt the “branch rollout” strategy from the original MBPO algorithm: for each interaction step in the simulation, multiple $k$-step rollouts are performed in the latent space to augment the dataset for policy learning. Finally, during the policy optimization phase, we use the standard SAC algorithm.

Overall, our approach achieves bidirectional feedback between the simulation and the real environment through the design of the latent space. The real environment provides guidance for transforming observations in the simulation during the shared latent space training process, while the simulation, mediated by the shared latent space, enhances the modeling of the real environment. Through the joint learning of $p^{\text{sim}}$ and $p^{\text{real}}$, the model is capable of identifying both the differences in identical observations and the dynamic relationships between different observations across the two environments. This enables the simulation to expand the coverage of the real environment’s limited offline data via interactive exploration. Moreover, initializing the latent space environment model using a single environment helps reduce the overall learning difficulty and enhances the feasibility of mutual guidance between the two environments.

\section{Experiments}

In our experiments, we aim to investigate the following three questions: (1) What is the impact of the sim-to-real gap on the effectiveness of policy transfer? (2) How do sim-to-real gap affect model-based RL? (3) To what extent can model-based methods accurately perceive the sim-to-real gap?

Our experiments aim to reveal the difficulty of model-based RL facing the sim-to-real gap through our proposed method from multiple perspectives. During the experiment, we used the HalfCheetah-v2 environment of MuJoCo~\cite{todorov2012mujoco} and distinguished between simulation and real environment by changing environmental parameters (such as gravity acceleration, torso length, and thigh length).

\subsection{Evaluation of Direct Policy Transfer}

\begin{figure} 
    \centering
    \includegraphics[width=1\linewidth]{  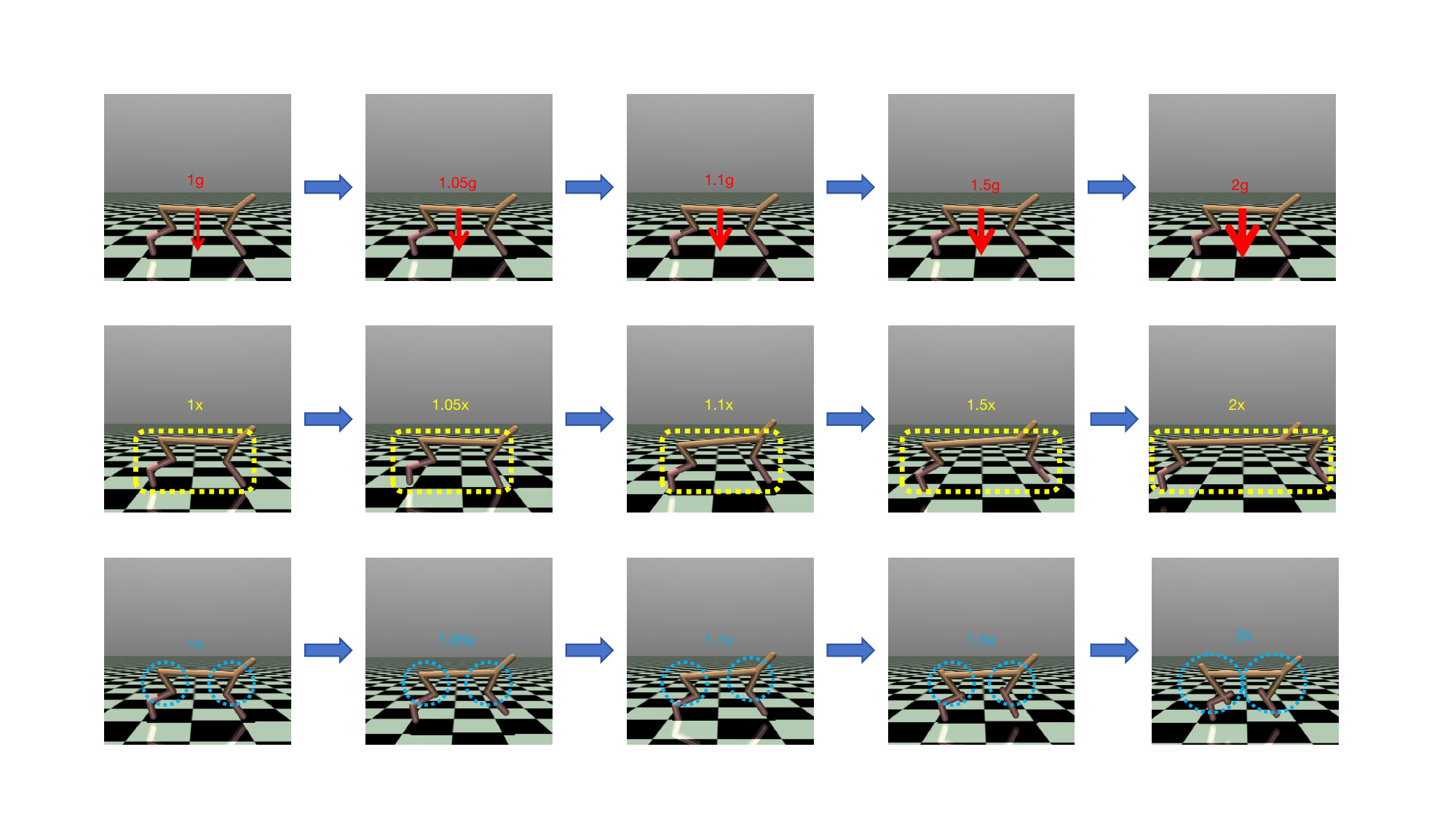}
    \caption{HalfCheetah-v2 with different dynamic perturbations. The first row changes the \textbf{gravity acceleration}, the second row changes the \textbf{torso length} (and the corresponding leg position), and the third row changes the \textbf{thigh length}.}
    \label{fig:per}
\end{figure}

To study question (1), we use the MBPO algorithm to train the policy in the standard HalfCheetah-v2 environment and directly apply it to the environment with different perturbations to observe the average return. We used three different methods of perturbations: gravity acceleration, torso length, and thigh length. Each perturbation method contains different degrees of perturbation, as shown in Figure \ref{fig:per}.

Our experimental results are shown in Table \ref{tab:1}. It can be seen that under the perturbations of three different methods, the policy trained in the standard environment will show a significant performance degradation when used in the perturbation environment, and the degree of degradation is closely related to the dynamic gap between the environments.
This tells us that when the sim-to-real gap is large, we cannot expect that the policy learned in simulation can be directly transferred to the real environment to have a better performance. In addition, different environmental perturbations have different effects on the performance of policy transfer. Compared with the perturbations to torso length and thigh length, the perturbation to gravity acceleration has a more significant impact on policy performance, which shows that the HalfCheetah-v2 agent is more sensitive to changes in the gravity environment, and it is not easy to see directly from the visual information of the agent. Therefore, how to measure the sim-to-real gap and alleviate its negative effects on policy performance is a key research topic.

\begin{table}[htbp]
\centering
\caption{Average return of the policy trained in the standard environment in each perturbation environment}
\begin{tabular}{l|ccccc}
\toprule
Environment & 1$\times$ & 1.05$\times$ & 1.1$\times$ & 1.5$\times$ & 2$\times$ \\
\midrule
\textcolor[HTML]{C00000}{Gravity Acceleration} & 11810 & 11302 & 10889 & 4895 & 1693 \\
\textcolor[HTML]{D1B762}{Torso Length}  & 11810 & 11474 & 10544 & 5924 & 4802 \\
\textcolor[HTML]{0070C0}{Thigh Length}   & 11810 & 11607 & 9930 & 8037  & 4006 \\
\bottomrule
\end{tabular}
\label{tab:1}
\end{table}

\subsection{Latent Space Based Method for Mitigating the Sim-to-Real Gap}
To study question (2), we test the ability of our latent space based method to mitigate the sim-to-real gap. Here we regard the HalfCheetah-v2 environment under various perturbations as simulation, and the standard environment as the real environment. We follow the H2O settings, allowing agents to interact without restrictions in simulation and only using offline data in the real environment. We test whether our latent space based method can achieve better results when facing the sim-to-real gap compared to directly training the policy through the MBPO algorithm using simulation and real environment data.

For the offline dataset in the real environment, we used the \texttt{halfcheetah-medium-replay-v2} and \texttt{halfcheetah-medium-v2} datasets in D4RL~\cite{fu2020d4rl}. The former contains the data in the replay buffer during the training of a medium-quality agent in the HalfCheetah-v2 environment using the SAC algorithm, while the latter contains the data collected using this medium-quality agent.

During the training process, in both latent space based and non latent space based methods, we first use offline data in the real environment to train the single environment model and policy through the standard MBPO algorithm, then use this model to initialize the latent space model, and use this policy to initialize the policy in the latent space. The purpose of this is to provide a better starting point for learning the environment model and policy in the latent space, allowing the model to initially explore the information in the offline data, and prevent the information provided by the offline data from being overwhelmed by the information provided by the data obtained through unlimited interaction in simulation during interactive training.

To evaluate the performance of our method and test its ability to use both simulation and real environment data, we consider calculating the sum of the average returns of a policy in the two environments and use this to select the best policy during training for comparison. Our results are presented in Table \ref{tab:medium-replay-v2-results} and Table \ref{tab:medium-v2-results}.

\begin{table}[htbp]
\centering
\caption{Performance comparison on \texttt{halfcheetah-medium-replay-v2}, where \textbf{Offline Only} shows the results obtained by training with the standard MBPO algorithm only on the \texttt{halfcheetah-medium-replay-v2} dataset.}
\label{tab:medium-replay-v2-results}
\begin{tabular}{llccc|ccc}
\toprule
\multirow{2}{*}{\textbf{Environment}} & \multirow{2}{*}{\textbf{Scale}} & 
\multicolumn{3}{c|}{\textbf{w/ Latent Space}} & \multicolumn{3}{c}{\textbf{w/o Latent Space}} \\
 & & Sim & Real & \textbf{Sim+Real} & Sim & Real & \textbf{Sim+Real} \\
 \midrule
 \textbf{Offline Only}& -- & -- & 8376 & -- & -- & 8376 & -- \\
 \midrule
\multirow{4}{*}{\textcolor[HTML]{C00000}{Gravity Acceleration}} 
    & 1.05$\times$ & 9786 & 9867 & 19653 & 10364 & 10967 & \textbf{21331} \\
    & 1.1$\times$  & 9509 & 9801 & 19310 & 9676 & 10005 & \textbf{19681} \\
    & 1.5$\times$  & 7428 & 7477 & \textbf{14905} & 7220 & 7027 & 14247 \\
    & 2$\times$    & 4476 & 6060 & \textbf{10536} & 5297 & 5228 & 10525 \\
\midrule
\multirow{4}{*}{\textcolor[HTML]{D1B762}{Torso Length}} 
    & 1.05$\times$ & 9586 & 9618 & 19204 & 10129 & 10284 & \textbf{20413} \\
    & 1.1$\times$  & 9106 & 9362 & 18468 & 10273 & 10235 & \textbf{20508} \\
    & 1.5$\times$  & 7411 & 7351 & 14762 & 8899 & 7724 & \textbf{16623} \\
    & 2$\$times$    & 5985 & 7492 & \textbf{13477} & 5899 & 7161 & 13060 \\
\midrule
\multirow{4}{*}{\textcolor[HTML]{0070C0}{Thigh Length} } 
    & 1.05$\times$ & 9931 & 10069 & 20000 & 10664 & 10593 & \textbf{21257} \\
    & 1.1$\times$  & 9906 & 9496 & 19402 & 10967 & 10263 & \textbf{21230} \\
    & 1.5$\times$  & 8249 & 6012 & 14261 & 9168 & 6875 & \textbf{16043} \\
    & 2$\times$    & 8319 & 4784 & 13103 & 8652 & 4972 & \textbf{13624} \\
\bottomrule
\end{tabular}
\end{table}

\begin{table}[htbp]
\centering
\caption{Performance comparison on \texttt{halfcheetah-medium-v2}, where \textbf{Offline Only} shows the results obtained by training with the standard MBPO algorithm only on the \texttt{halfcheetah-medium-v2} dataset.}
\label{tab:medium-v2-results}
\begin{tabular}{llccc|ccc}
\toprule
\multirow{2}{*}{\textbf{Environment}} & \multirow{2}{*}{\textbf{Scale}} & 
\multicolumn{3}{c|}{\textbf{w/ Latent Space}} & \multicolumn{3}{c}{\textbf{w/o Latent Space}} \\
 & & Sim & Real & \textbf{Sim+Real} & Sim & Real & \textbf{Sim+Real} \\
 \midrule
 \textbf{Offline Only}& -- & -- & 9407 & -- & -- & 9407 & -- \\
\midrule
\multirow{4}{*}{\textcolor[HTML]{C00000}{Gravity Acceleration}} 
    & 1.05$\times$ & 9792 & 9998 & 19790 & 10194 & 10597 & \textbf{20791} \\
    & 1.1$\times$  & 9019 & 9333 & 18352 & 9078 & 9587 & \textbf{18665} \\
    & 1.5$\times$  & 7343 & 7257 & 14600 & 7750 & 7358 & \textbf{15108} \\
    & 2$\times$    & 5151 & 6141 & \textbf{11292} & 6250 & 5038 & 11288 \\
\midrule
\multirow{4}{*}{\textcolor[HTML]{D1B762}{Torso Length}} 
    & 1.05$\times$ & 9614 & 9833 & 19447 & 10191 & 10236 & \textbf{20427} \\
    & 1.1$\times$  & 9540 & 9680 & 19220 & 10432 & 10208 & \textbf{20640} \\
    & 1.5$\times$  & 7408 & 7414 & \textbf{14822} & 7047 & 7707 & 14754 \\
    & 2$\times$    & 6074 & 7240 & \textbf{13314} & 6043 & 7190 & 13233 \\
\midrule
\multirow{4}{*}{\textcolor[HTML]{0070C0}{Thigh Length} } 
    & 1.05$\times$ & 9581 & 9803 & 19384 & 10629 & 10573 & \textbf{21202} \\
    & 1.1$\times$  & 10413 & 9967 & 20380 & 10804 & 10805 & \textbf{21609} \\
    & 1.5$\times$  & 8057 & 5991 & \textbf{14048} & 8319 & 5241 & 13560 \\
    & 2$\times$    & 8351 & 4472 & 12823 & 8272 & 5226 & \textbf{13498} \\
\bottomrule
\end{tabular}
\end{table}

The following facts can be observed from the experimental results. 

\textbf{First, the addition of the simulation may not necessarily have a positive effect on the policy training in the real environment.} Compared with using only offline samples in the real environment, the simulation can supplement more samples for training around the current policy, but due to the imperfection of the simulation, the addition of these samples may have a negative effect on policy training. From Tables \ref{tab:medium-replay-v2-results} and \ref{tab:medium-v2-results}, it can be seen that under the two settings of using and not using latent space, in the environment where various perturbation methods are applied, a small degree of perturbation (1.05$\times$, 1.1$\times$) on the original environment parameters can usually improve the performance of the policy in the real environment, while a large degree of perturbation (1.5$\times$, 2$\times$) will reduce the performance of the policy in the real environment. 

\textbf{Second, the effect of using latent space methods to mitigate the sim-to-real gap is related to the degree of environmental perturbation.} Under a smaller degree of perturbation (1.05x, 1.1x), the method using latent space performs better than the method not using latent space, while under a larger degree of perturbation (1.5x, 2x), the method using latent space will narrow the performance gap with another and often achieve higher performance. By mapping the observations in simulation and real environment into a common latent space, the samples in the two environments can be better distinguished. Therefore, the method using latent space reduces the mutual interference between samples in the two environments under a larger degree of environmental perturbation. Under a smaller degree of environmental perturbation, simulation can directly play a positive role in supplementing the real environment, and the use of latent space increases the difficulty of training, which leads to poor performance.

\textbf{Third, simulations formed by different perturbations have different effects on the real environment.} In Tables \ref{tab:medium-replay-v2-results} and \ref{tab:medium-v2-results}, under the perturbations of gravity acceleration and torso length, the performance of the policy in simulation and real environment will decrease synchronously as the degree of perturbation increases. However, under the perturbation of thigh length, the policy seems to be more inclined to maintain and improve the performance in simulation, ignoring the maintenance of performance in the real environment. We speculate that this may be because the dynamic characteristics of the same observations in simulation and real environment are too different at this time, making it difficult to find a simple correlation between the observations of the two environments. Therefore, during the training process of the environment model and strategy, the offline samples in the real environment are overwhelmed by the interactive samples that are constantly supplemented in the simulation, resulting in the policy's preference for the performance in simulation. This may also be the reason why the method using latent space performs poorly in such environments.

\subsection{Latent Space Based Method for Quantifying the Sim-to-Real Gap}

To study question (3), we examine the change in the cross-domain mapping $m$ after training our latent space based method. When the latent space is initialized using the environment model trained with offline data, the dynamic gap between different environments has not yet been measured, and $m$ is initialized to the identity mapping. During the interactive training phase, the larger the difference in environmental dynamics between the simulation and the real environment, the farther the distance between the same observations in the latent space should be, and the more $m$ should be different from the identity mapping. Therefore, we compare the distance between the trained $m$ and the identity mapping to observe its relationship with the sim-to-real gap. Specifically, for the observation $o$ in the offline dataset, we calculate the Euclidean distance between $m \circ p^{\text{real}}(o)$ and $p^{\text{real}}(o)$, and then we average it over all $o$.

\begin{table}[htbp]
\centering
\caption{Average Euclidean distance between $m \circ p^{\text{real}}(o)$ and $p^{\text{real}}(o)$ for \texttt{halfcheetah-medium-replay-v2}}
\begin{tabular}{l|ccccc}
\toprule
Environment  & 1.05$\times$ & 1.1$\times$ & 1.5$\times$ & 2$\times$ \\
\midrule
\textcolor[HTML]{C00000}{Gravity Acceleration}  & 0.2189 & 0.3168 & 4.5248 & 0.1115 \\
\textcolor[HTML]{D1B762}{Torso Length} & 0.0519 & 0.1074 & 3.5134 & 20.8568 \\
\textcolor[HTML]{0070C0}{Thigh Length}   & 0.0418 & 0.0188 & 1.1804  & 6.2654 \\
\bottomrule
\end{tabular}
\label{tab:4}
\end{table}

\begin{table}[htbp]
\centering
\caption{Average Euclidean distance between $m \circ p^{\text{real}}(o)$ and $p^{\text{real}}(o)$ for \texttt{halfcheetah-medium-v2}}
\begin{tabular}{l|ccccc}
\toprule
Environment  & 1.05$\times$ & 1.1$\times$ & 1.5$\times$ & 2$\times$ \\
\midrule
\textcolor[HTML]{C00000}{Gravity Acceleration} & 0.3589 & 3.5410 & 6.7007 & 20.7380 \\
\textcolor[HTML]{D1B762}{Torso Length}   & 3.0851 & 2.6206 & 9.5777 & 30.8118 \\
\textcolor[HTML]{0070C0}{Thigh Length}    & 13.3575 & 0.7880 & 247.8369  & 385.0432 \\
\bottomrule
\end{tabular}
\label{tab:5}
\end{table}

Tables \ref{tab:4} and \ref{tab:5} show the experimental results of the \texttt{halfcheetah-medium-replay-v2} and \texttt{halfcheetah-medium-v2} datasets respectively. The following information can be obtained from the tables.

\textbf{First, the latent space based method has a certain ability to identify and measure the sim-to-real gap.} Under each environmental perturbation method, as the degree of perturbation increases, the gap between $m$ and the identity mapping generally shows a trend of gradual increase, which shows that the latent space based method can measure the sim-to-real gap to a certain extent. 

\textbf{Second, the use of different offline data will cause $m$ to have different performances during training.} Compared with the \texttt{halfcheetah-medium-replay-v2} dataset, the gap between $m$ and the identity mapping under \texttt{halfcheetah-medium-v2} training is larger, especially in the \textcolor[HTML]{0070C0}{Thigh Length} perturbation environment. We speculate that in addition to the influence of the number of offline samples, this is because the samples in the latter have a higher intersection with the policy optimization direction in simulation, thus forcing the latent space to make a clearer distinction between the same observations in the two environments, while the former offline samples themselves have little correlation with the policy optimization direction in simulation, so there will not be a significant degree of interference in the training of environmental modeling, which curbs the learning of $m$.

In general, this experiment shows that the effect of the latent space based method on the sim-to-real gap is closely related to the distribution of offline samples in the real environment. This further shows that the learning effect of the policy of the model-based method in the sim-to-real gap situation is affected by the actual sample distribution of the real environment, and the final training performance cannot be judged solely by the degree of environmental disturbance.

\section{Challenges}

In this section, we will analyze and summarize the challenges encountered by model-based methods in sim-to-real migration tasks based on our experimental results.

\subsection{Imperfection of the Environment Model}

Due to the limited representation ability of the environment model, it can only perform well near the distribution of training data, which leads to greater risks in the sim-to-real transfer scenario. On the one hand, using imperfect models trained with simulated data to initialize the dynamic prediction mechanism in the real environment may aggravate the negative transfer phenomenon~\cite{tan2018survey}. On the other hand, we found that the model is "locally good", that is, the model can only accurately predict near the sample distribution in the current replay buffer, and the error in the area outside the distribution is larger than that of the random model. Since the dynamic changes of the environment usually lead to changes in the state distribution of policy dependence, this phenomenon significantly increases the risk in model transfer—when transferring to a new environment, the new environment may not utilize areas where the old environment model performs well, but instead explore and train in areas where it performs poorly, causing the difficulty of adapting to the new environment to increase rather than decrease.

\subsection{Latent Representation Shift across Sim-to-Real Transfer}

The second challenge encountered by model-based sim-to-real transfer is the shift in latent representation between the simulation environment and the real environment. Although the simulation and real environment provide observations in the same form, such as images, location information, sensor data, etc., they actually reflect two slightly different real dynamic processes. When we adopt a latent state based modeling approach, we can see this difference more clearly: even if the observations are numerically close, their representations in latent space may be far apart.
This difference shows that the context of the same observation in the simulation environment (such as physical dynamics or future evolution trends) is not consistent with the context in the real environment. In other words, the meaning of "representation" in simulation and real environment may be completely different for inputs that look the same. This shift in latent representation will cause the decision-making basis of the policy in latent space to fail, thereby weakening the generalization ability in the real environment.
This latent shift not only poses a challenge to policy transfer, but also means that attempts to build a unified environment model in model-based methods may be fundamentally limited. Only by recognizing the implicit mechanism behind the semantic differences of such observations can we design more robust sim-to-real transfer methods.

\subsection{Hard Information Exchange between Simulation and Real Environment}

Through our latent space based approach, we found that there are challenges in information exchange between the simulation and the real environment. Through joint learning of samples from the two environments, the model often finds it difficult to discover the association between samples from different environments (such as the dynamic equivalence of normal posture in a standard environment and low-center-of-gravity posture in a high-gravity environment during walking), but tends to separate samples from the two environments. Table \ref{tab:6} shows the relationship between the KL divergence in the latent space and the KL divergence in the original space when training with \texttt{halfcheetah-medium-v2}, where the samples in the real environment and simulation use offline data and the replay buffer during training, respectively. As can be seen from the table, the samples in the latent space are more scattered than those in the original space, and the values are highly consistent with those in Table \ref{tab:5}. This shows that the latent space can only discover the gap between the same observations in the two environments, but it is difficult to discover the association between different observations. This further shows the difficulty of samples in simulation to provide additional valuable information to the real environment, and the model-based method has a tendency to distinguish rather than utilize data in simulation.

\begin{table}[htbp]
\centering
\caption{Ratio of latent-space to original-space KL divergence between real environment \texttt{halfcheetah-medium-v2} samples and simulation replay buffer samples during training. Original-space KL divergence is \textbf{14.2134} on average.}
\begin{tabular}{l|ccccc}
\toprule
Environment  & 1.05$\times$ & 1.1$\times$ & 1.5$\times$ & 2$\times$ \\
\midrule
\textcolor[HTML]{C00000}{Gravity Acceleration} & 31.1056 & 240.3639 & 233.6854 & 1649.0747 \\
\textcolor[HTML]{D1B762}{Torso Length}   & 196.8728 & 162.0648 & 206.4534 & 636.1692 \\
\textcolor[HTML]{0070C0}{Thigh Length}    & 934.9126 & 161.6329 & 42751.3436  & 55422.8860 \\
\bottomrule
\end{tabular}
\label{tab:6}
\end{table}

\section{Limitations}

Our work has the following limitations. First, to analyze the challenges faced by model-based approaches in sim-to-real transfer, we proposed a latent space based method. However, our experiments were primarily conducted on MBPO, a representative model-based algorithm, and did not cover a broader range of approaches such as MOPO~\cite{yu2020mopo} or PETS~\cite{chua2018deep}, which may exhibit different behaviors when dealing with the sim-to-real gap. Second, all our experiments were conducted in the MuJoCo environment, where the distinction between the simulation and the real environment was made by varying environment parameters. We did not perform evaluations in the real physical world, which limits our ability to directly observe the method's performance in real-world scenarios. Moreover, all experiments involving latent representations were limited to the HalfCheetah-v2 environment, which constrains the generalizability of our findings.

\section{Conclusion}

In this work, we propose a latent space based analytical framework to systematically evaluate the performance of model-based reinforcement learning (MBRL) methods in sim-to-real transfer tasks. The core idea of our method is to construct a shared latent representation space between the simulation and the real environment. Through the design of structured loss functions, the latent space is encouraged to capture the essential and invariant dynamic properties across environments. This latent representation aims to transform the underlying partially observable Markov decision process (POMDP) into a more tractable fully observable MDP, thereby facilitating more stable and transferable policy learning.

We argue that the latent space based approach represents a natural extension to existing model-based methods, offering both the capacity to mitigate the impact of the sim-to-real gap and the ability to quantify the extent of this gap in the learned representations. Preliminary experiments conducted in the MuJoCo environment confirm that the sim-to-real gap can lead to significant performance degradation in policy transfer. Moreover, our analysis reveals and summarizes several critical challenges faced by model-based approaches in sim-to-real scenarios, such as limited model generalization and difficulties in learning consistent dynamics across domains. These insights highlight important directions for future work.

\bibliographystyle{unsrt}  
\bibliography{references}  






\end{document}